\ifdefined\XeTeXversion\else\pdfoutput=1\fi
\documentclass[11pt]{article}
\usepackage[final]{preprint}

\newcommand{\numPlanted}{424}
\newcommand{\numScoreAstra}{74.8}
\newcommand{\numScoreArgon}{61.3}
\newcommand{\numScoreOpus}{53.6}
\newcommand{\numScoreFable}{52.2}
\newcommand{\numScoreSol}{46.6}
\newcommand{\numScoreGrok}{43.4}
\newcommand{\numScoreWonder}{39.6}
\newcommand{\numScoreSonnet}{28.6}
\newcommand{\numScoreHaiku}{10.4}
\newcommand{\numScoreLuna}{17.1}

\newcommand{\numFoundAstra}{79.0}
\newcommand{\numFoundArgon}{66.5}
\newcommand{\numFoundOpus}{61.0}
\newcommand{\numFoundFable}{59.1}
\newcommand{\numFoundSol}{51.4}
\newcommand{\numFoundGrok}{48.9}
\newcommand{\numFoundWonder}{45.5}
\newcommand{\numBestUnfixed}{25.2}

\newcommand{\numReadPastMin}{20.2}
\newcommand{\numReadPastMax}{51.3}

\newcommand{\numNotReadGrokPct}{7.4}
\newcommand{\numNotReadMaxPct}{1.3}

\newcommand{\numRepetitionPlanted}{24}

\newcommand{\numEmojiPlanted}{13}

\newcommand{\numConnectiveMax}{42.0}

\newcommand{\numRepetitionOthersMaxPct}{37.5}

\newcommand{\numFamFixedFactualAstra}{82.6}
\newcommand{\numFamReadPastFactualAstra}{12.3}
\newcommand{\numFamFixedFactualArgon}{49.2}
\newcommand{\numFamReadPastFactualArgon}{45.2}
\newcommand{\numFamFixedFactualOpus}{50.0}
\newcommand{\numFamReadPastFactualOpus}{29.4}
\newcommand{\numFamFixedFactualFable}{47.6}
\newcommand{\numFamReadPastFactualFable}{33.8}
\newcommand{\numFamFixedFactualSol}{54.8}
\newcommand{\numFamReadPastFactualSol}{40.0}
\newcommand{\numFamFixedFactualGrok}{52.3}
\newcommand{\numFamReadPastFactualGrok}{26.9}
\newcommand{\numFamFixedFactualWonder}{45.4}
\newcommand{\numFamReadPastFactualWonder}{42.4}
\newcommand{\numFamReadPastFactualAll}{32.9}

\newcommand{\numFamFixedConflictAstra}{77.1}
\newcommand{\numFamReadPastConflictAstra}{7.5}
\newcommand{\numFamFixedConflictArgon}{63.0}
\newcommand{\numFamReadPastConflictArgon}{10.2}
\newcommand{\numFamFixedConflictOpus}{61.7}
\newcommand{\numFamReadPastConflictOpus}{5.6}
\newcommand{\numFamFixedConflictFable}{60.4}
\newcommand{\numFamReadPastConflictFable}{10.3}
\newcommand{\numFamFixedConflictSol}{61.4}
\newcommand{\numFamReadPastConflictSol}{16.5}
\newcommand{\numFamFixedConflictGrok}{70.3}
\newcommand{\numFamReadPastConflictGrok}{10.7}
\newcommand{\numFamFixedConflictWonder}{57.3}
\newcommand{\numFamReadPastConflictWonder}{18.7}
\newcommand{\numFamReadPastConflictAll}{11.4}

\newcommand{\numFamFixedEditorialAstra}{69.9}
\newcommand{\numFamReadPastEditorialAstra}{29.3}
\newcommand{\numFamFixedEditorialArgon}{53.8}
\newcommand{\numFamReadPastEditorialArgon}{45.0}
\newcommand{\numFamFixedEditorialOpus}{53.2}
\newcommand{\numFamReadPastEditorialOpus}{43.4}
\newcommand{\numFamFixedEditorialFable}{52.4}
\newcommand{\numFamReadPastEditorialFable}{45.0}
\newcommand{\numFamFixedEditorialSol}{33.3}
\newcommand{\numFamReadPastEditorialSol}{66.0}
\newcommand{\numFamFixedEditorialGrok}{28.8}
\newcommand{\numFamReadPastEditorialGrok}{62.8}
\newcommand{\numFamFixedEditorialWonder}{25.0}
\newcommand{\numFamReadPastEditorialWonder}{73.5}
\newcommand{\numFamReadPastEditorialAll}{52.1}

\newcommand{\numFamFixedFormattingAstra}{59.5}
\newcommand{\numFamReadPastFormattingAstra}{37.9}
\newcommand{\numFamFixedFormattingArgon}{66.3}
\newcommand{\numFamReadPastFormattingArgon}{29.5}
\newcommand{\numFamFixedFormattingOpus}{28.4}
\newcommand{\numFamReadPastFormattingOpus}{68.9}
\newcommand{\numFamFixedFormattingFable}{33.7}
\newcommand{\numFamReadPastFormattingFable}{61.1}
\newcommand{\numFamFixedFormattingSol}{32.2}
\newcommand{\numFamReadPastFormattingSol}{64.4}
\newcommand{\numFamFixedFormattingGrok}{22.7}
\newcommand{\numFamReadPastFormattingGrok}{66.3}
\newcommand{\numFamFixedFormattingWonder}{22.7}
\newcommand{\numFamReadPastFormattingWonder}{73.9}
\newcommand{\numFamReadPastFormattingAll}{57.4}

\newcommand{\numFamFixedVoiceAstra}{80.7}
\newcommand{\numFamReadPastVoiceAstra}{16.7}
\newcommand{\numFamFixedVoiceArgon}{70.1}
\newcommand{\numFamReadPastVoiceArgon}{26.5}
\newcommand{\numFamFixedVoiceOpus}{73.1}
\newcommand{\numFamReadPastVoiceOpus}{18.9}
\newcommand{\numFamFixedVoiceFable}{64.8}
\newcommand{\numFamReadPastVoiceFable}{21.2}
\newcommand{\numFamFixedVoiceSol}{51.1}
\newcommand{\numFamReadPastVoiceSol}{42.4}
\newcommand{\numFamFixedVoiceGrok}{42.8}
\newcommand{\numFamReadPastVoiceGrok}{43.9}
\newcommand{\numFamFixedVoiceWonder}{47.0}
\newcommand{\numFamReadPastVoiceWonder}{48.1}
\newcommand{\numFamReadPastVoiceAll}{31.1}

\newcommand{\numDemotedPlanted}{5}
\newcommand{\numWebFactsPlanted}{83}

\newcommand{\numAuditCostMin}{1.78}
\newcommand{\numAuditCostMax}{8.61}

\newcommand{\numRealErrors}{175}

\newcommand{\numScoreMeanFrontier}{53.1}
\newcommand{\numRealErrorPages}{103}
\newcommand{\numBrands}{11}

\newcommand{\numGradingCost}{0.62}

\newcommand{\numCheatCallsMin}{41}
\newcommand{\numCheatCallsMax}{132}

\newcommand{\numCheatCallsThrough}{7}

\newcommand{\numCheatBlockedPct}{98.7}

\newcommand{\numRawAgreement}{77.7}
\newcommand{\numACOne}{0.72}

\newcommand{\numPerFamilyTarget}{8}
\newcommand{\numFamFactual}{83}

\newcommand{\numConflictMatching}{63}
\newcommand{\numConflictReasoning}{20}
\newcommand{\numKindDangling}{17}
\newcommand{\numKindIrrelevant}{23}
\newcommand{\numKindConnective}{23}
\newcommand{\numKindHeadingBody}{31}
\newcommand{\numKindListItem}{25}
\newcommand{\numKindListFlat}{12}
\newcommand{\numKindCell}{10}
\newcommand{\numKindSlopParaphrase}{20}
\newcommand{\numKindSlopScaffolding}{10}
\newcommand{\numKindUrgency}{13}
\newcommand{\numKindAttitude}{12}
\newcommand{\numKindAudience}{11}
\newcommand{\numKindLegalese}{9}

\newcommand{\numFactSourceHosts}{85}

\newcommand{\numPagesPerBrand}{25}
\newcommand{\numMaxTurns}{48}
\newcommand{\numMaxToolCalls}{200}
\newcommand{\numMaxSearches}{20}
\newcommand{\numMaxMinutes}{30}
\newcommand{\numMaxInputTokens}{5M}
\newcommand{\numMaxOutputTokens}{100K}
\newcommand{\numResponseTokens}{32K}
\newcommand{\numContextTokens}{360K}
\newcommand{\numCallSeconds}{300}
\newcommand{\numCallAttempts}{3}
\newcommand{\numSearchResults}{10}
\newcommand{\numSearchPrice}{0.005}
\newcommand{\numWebArchives}{14}
\newcommand{\numCopyPassages}{3}
\newcommand{\numCopyChars}{80}

\newcommand{\numAuditCostOpus}{3.16}
\newcommand{\numAuditCostFable}{7.61}
\newcommand{\numAuditCostSol}{1.82}
\newcommand{\numAuditCostGrok}{2.89}
\newcommand{\numAuditCostWonder}{2.42}

\newcommand{\numSearchMultiHopPct}{0.6}
\newcommand{\numAuditCostHaiku}{0.08}
\newcommand{\numAuditCostLuna}{0.15}
\newcommand{\numAuditCostSonnet}{0.98}

\newcommand{\numWebFactsLostPctMin}{14.9}
\newcommand{\numWebFactsLostPctMax}{19.8}
\newcommand{\numWebCorrectedOnOpus}{50.0}
\newcommand{\numWebCorrectedOffOpus}{33.7}
\newcommand{\numWebCorrectedOnSol}{54.8}
\newcommand{\numWebCorrectedOffSol}{35.0}
\newcommand{\numWebCorrectedOnArgon}{49.2}
\newcommand{\numWebCorrectedOffArgon}{34.3}
\newcommand{\numWebRightOnOpus}{64.8}
\newcommand{\numWebRightOffOpus}{63.5}
\newcommand{\numWebRightOnSol}{58.0}
\newcommand{\numWebRightOffSol}{52.2}
\newcommand{\numWebRightOnArgon}{53.1}
\newcommand{\numWebRightOffArgon}{52.6}

\newcommand{\numEffortDetGainSol}{4.7}
\newcommand{\numEffortCorrGainSol}{6.9}
\newcommand{\numEffortDetGainOpus}{5.1}
\newcommand{\numEffortCorrGainOpus}{4.8}
\newcommand{\numWebCorrectedDropOpus}{16.3}
\newcommand{\numWebCorrectedDropSol}{19.8}
\newcommand{\numWebCorrectedDropArgon}{14.9}
\newcommand{\numWebRightDropOpus}{1.4}
\newcommand{\numWebRightDropSol}{5.8}
\newcommand{\numWebRightDropArgon}{0.5}
\newcommand{\numWebRightGapOnOpus}{14.9}
\newcommand{\numWebRightGapOnSol}{3.1}
\newcommand{\numWebRightGapOnArgon}{3.9}
\newcommand{\numWebRightRunGapOffMax}{7.5}

\newcommand{\numSearchKnewPct}{40.7}
\newcommand{\numSearchNoWebPct}{38.7}

\newcommand{\numSearchFixes}{940}
\newcommand{\numSearchOneHopPct}{15.0}

\newcommand{\numSearchUnclearPct}{4.9}
\newcommand{\numSearchHandCheckPct}{86.7}

\newcommand{\numHumanFixedAgree}{70.0}
\newcommand{\numHumanWeakAgree}{0.0}
\newcommand{\numHumanWrongAgree}{20.0}
\newcommand{\numHumanMissedAgree}{100.0}
\newcommand{\numHumanAcceptedAgree}{80.0}
\newcommand{\numHumanRejectedAgree}{37.5}
\newcommand{\numHumanDetectAgree}{90.0}

\newcommand{\numWeakShare}{10}

\newcommand{\numPersRunsRead}{231}

\newcommand{\numPersRepeatSol}{47.0}
\newcommand{\numPersFilingWebAstra}{54}
\newcommand{\numPersFilingWebOthersMin}{2}
\newcommand{\numPersFilingWebOthersMax}{22}

\newcommand{\numPersNaturalHeldAstra}{91.9}
\newcommand{\numPersNaturalHeldOthers}{42.6}

\newcommand{\numPersWebBeforeFirstArgon}{84}

\newcommand{\numPersDiagBlockArgon}{98}
\newcommand{\numPersDiagBlockOthersMax}{5}
\newcommand{\numPersFiledByCheckpointOpus}{62}
\newcommand{\numPersProbeFable}{97.0}
\newcommand{\numPersTypedUrlFable}{76.2}
\newcommand{\numPersFilingNoReadSol}{98}

\newcommand{\numPersOrEmptyGrok}{92.0}
\newcommand{\numPersLimitEndGrok}{60.6}

\newcommand{\numPersHtmlWonder}{28}
\newcommand{\numPersHtmlArgon}{75}

\newcommand{\numJevMatchKappa}{0.89}

\usepackage{amsmath,amsfonts}
\usepackage{colortbl}
\usepackage{booktabs,arydshln}
\usepackage{multirow}
\usepackage[normalem]{ulem}
\usepackage{pifont}
\usepackage{relsize}
\usepackage{enumitem}
\usepackage{graphicx}
\usepackage{placeins}

\usepackage{booktabs}
\usepackage{array}
\definecolor{errorTint}{HTML}{FBE9E7}
\definecolor{errorInk}{HTML}{B3412B}
\definecolor{guideTint}{HTML}{E8F0FB}
\definecolor{guideInk}{HTML}{2F5D9E}
\definecolor{ruleGray}{gray}{0.75}
\definecolor{tabMuted}{HTML}{6B7785}
\newcommand{\famname}[3]{\leavevmode{\color{#1}\textbf{#2}}\newline{\color{tabMuted}\footnotesize\textit{tests #3}}}
\newcommand{\kc}[1]{\,{\footnotesize(#1)}}
\newcommand{\kd}{~~$\cdot$~~}
\newcommand{\tierrow}[4]{\multicolumn{3}{>{\columncolor{#2}}l}{\rule[-0.9ex]{0pt}{3.2ex}{\color{#1}\textbf{\textsc{#3}}}\quad{\color{tabMuted}\footnotesize\textit{#4}}}\\}

\usepackage[colorinlistoftodos,textwidth=2.4cm,textsize=scriptsize]{todonotes}

\definecolor{claudeorange}{HTML}{D97757}

\newcommand{\marketbench}{\textsc{{Marketing-Bench}}}
\newcommand{\marketbenchtitle}{\textsc{\marketbench}}

\title{\marketbenchtitle{}: Evaluating Frontier Models and Agents on Real-World Marketing Capabilities}
\author{Sumanth Balaji \qquad Siddharth Gupta \qquad Ameet Deshpande \\
  \textbf{Vishvak Murahari} \qquad \textbf{Karthik Narasimhan} \\[4pt]
  \includegraphics[height=0.32in]{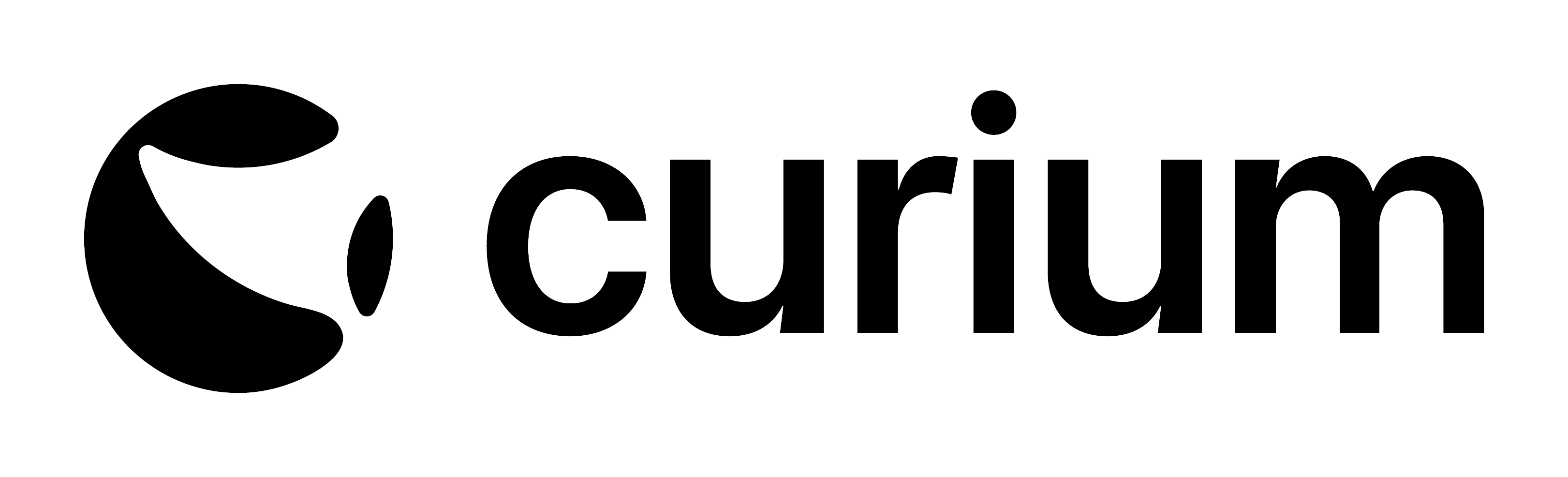}}
\date{}

\begin{document}
\maketitle

\begin{abstract}
As frontier AI models and agents are deployed more broadly in the real world, marketing is emerging as one of their largest applications.
Marketing teams use AI models to produce brand assets like website copy and social media posts, while consumers' agents autonomously discover products, research purchases, and navigate websites.
However, evaluating marketing capability is difficult because unlike areas like software development, customer support, or legal review where ground truth is more objectively defined, marketing is subjective, and output quality is judged based on taste and brand.
We introduce \marketbench{}, the first open-source marketing benchmark\footnote{\begin{tabular}[t]{@{}l@{\enspace}l@{}}Code: & \href{https://github.com/curium-ai/marketing-bench}{\nolinkurl{github.com/curium-ai/marketing-bench}}\\ Data: & \href{https://huggingface.co/datasets/CuriumAI/Marketing-Bench}{\nolinkurl{hf.co/datasets/CuriumAI/Marketing-Bench}}\end{tabular}}, which pairs discriminative tasks (detecting brand anomalies) with generative tasks (remediation), while keeping answers verifiable.
\marketbench{} tests five key pillars: factuality, consistency, coherence, structure, and brand voice, all crucial elements for good brand content on the web.
We find that frontier models are far from solving the benchmark: the average model corrects only about half of the introduced issues (\numScoreMeanFrontier\%), whereas a marketing expert corrects nearly all of them.
Even the strongest, GPT-6 Astra, reaches only \numScoreAstra\%, with Claude Opus and Fable, Gemini Argon, and GPT-6 Sol falling well short.
These findings suggest that marketing capabilities are less mature than adoption implies, and that principled evaluation should precede further scaling of usage.
\marketbench{} is a first step toward ``frontier marketing models'' with superhuman marketing performance.

\end{abstract}

\section{Introduction}
\label{sec:introduction}

\begin{figure*}[t]
  \centering
  \includegraphics[width=\textwidth]{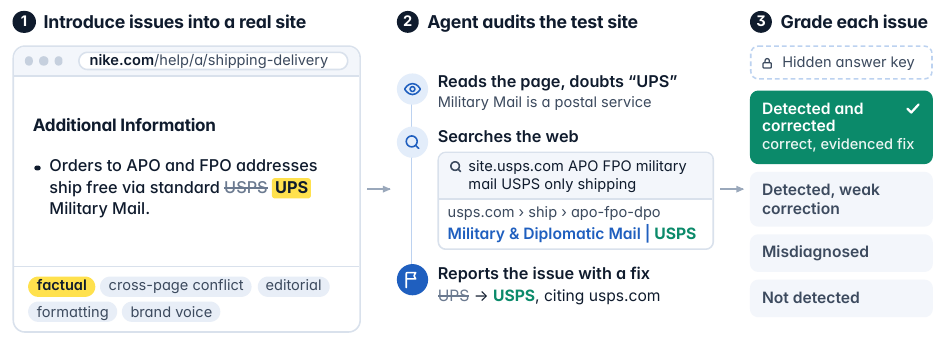}
  \caption{How \marketbench{} tests an agent. (1) We introduce issues from five families into a snapshot of a real brand site; here one word on a Nike help page changes from USPS to UPS. (2) The agent audits the test site with tools, checks the claim with a web search and reports the issue with a fix that cites usps.com. (3) A grader matches each finding to the hidden answer key and gives each introduced issue one of four outcomes; this one is detected and corrected.}
  \label{fig:hero}
\end{figure*}

Frontier AI models have moved from answering questions to doing work: they write and repair software, complete multi-step tasks on the web, and increasingly act on behalf of their users.
Marketing has become a major setting for this work, with frontier models now on both sides of it.
Brands use them to write the content they publish, from website copy and product pages to help articles and emails, while consumers send agents like ChatGPT, Muse and Instinct to research products and read that content on their behalf \citep{aggarwal2024geo}.
As these models are deployed by brands and consumers alike, we need a principled way to measure and improve their marketing capability.

Evaluating marketing ability, however, poses a distinct challenge.
Benchmarks in math, coding, web navigation and customer support each rest on an objective check, such as a final answer, a passing test suite or the end state of a website or database \citep{cobbe2021gsm8k,jimenez2024swebench,zhou2024webarena,yao2025taubench}.
Marketing is inherently subjective, as the quality of an output depends on the brand and the taste of its audience, thus making a single objective check hard to define.

To address this, we introduce \marketbench{}, to our knowledge the first open-source benchmark of the marketing ability of frontier models (\Cref{fig:hero}).
We frame the evaluation of marketing ability as a denoising task.
Denoising objectives corrupt a clean signal and ask a model to recover it, and they underpin much of modern representation learning and generative modeling \citep{vincent2008denoising,devlin2019bert,clark2020electra,lewis2020bart,raffel2020t5,ho2020denoising}.
We apply this recipe to brand content: we start from a brand's real pages, corrupt them with a small set of known edits, and ask an agent to restore them.
Restoring a page takes two steps: the agent must first detect which passages were corrupted, a discriminative step, and then correct each one, a generative step.

The clean pages come from real brand websites, spanning help centers, product documentation and pricing pages, and a generator model corrupts them with small, exact edits, each of which we call an \emph{introduced issue}.
These issues fall into five families in two tiers (\Cref{tab:taxonomy}): \emph{errors} (factual errors and cross-page conflicts) make a page state something incorrect, while \emph{guideline violations} (in editing, formatting and brand voice) leave the page correct but below editorial or brand standards.
A screening model checks every edit, and human annotators review every introduced issue before it enters an answer key with its location, a diagnosis, acceptable fixes and, for a factual error, an outside source that proves the correct value (\Cref{sec:benchmark}).

We evaluate seven frontier models from five labs in one shared agent harness, with three runs per brand (\Cref{sec:setup}).
We find that frontier agents are far from solving \marketbench{}: the best model, GPT-6 Astra, detects and corrects \numScoreAstra\% of the introduced issues, and the other six detect and correct between \numScoreWonder\% and \numScoreArgon\% (\Cref{sec:headline}).
Interestingly, most issues that go undetected sit in text the agent already read: six of the seven agents read nearly every introduced issue, yet they read past \numReadPastMin\% to \numReadPastMax\% of all introduced issues without detecting them (\Cref{sec:funnel}).

Our contributions are as follows: (1) We introduce \marketbench{}, the first open-source benchmark of marketing ability, built from verifiable tasks on real brand websites. (2) We construct a human-reviewed answer key of introduced issues across five families, each with its location, diagnosis and acceptable fixes. (3) We evaluate seven frontier models and find that none reliably detects and corrects the introduced issues, with most undetected issues sitting in text the model has already read.
We hope \marketbench{} enables the community to faithfully measure and push the frontier, as a first step towards frontier marketing models.

\begin{table*}[t]
  \centering
  \small
  \setlength{\aboverulesep}{0pt}\setlength{\belowrulesep}{0pt}
  \renewcommand{\arraystretch}{1.25}
  \begin{tabular}{@{\hspace{4pt}}>{\raggedright\arraybackslash}p{0.18\textwidth}>{\raggedright\arraybackslash}p{0.29\textwidth}>{\raggedright\arraybackslash}p{0.47\textwidth}@{}}
    \arrayrulecolor{guideInk}\toprule[1pt]
    \rule{0pt}{2.6ex}\textbf{Family} & \textbf{What the issue is} & \textbf{Kinds} {\footnotesize(number introduced)} \\[2pt]
    \midrule[0.6pt]
    \tierrow{errorInk}{errorTint}{Errors}{the page states something incorrect}
    \arrayrulecolor{ruleGray}
    \famname{errorInk}{Factual}{verification} & A claim on the page is false, and a source outside the brand's own domains establishes the correct value. & Wrong fact\kc{\numFamFactual} \\
    \cmidrule{1-3}
    \famname{errorInk}{Cross-page conflict}{consistency} & Two passages contradict each other about the same entity, date, region and audience. & Direct contradiction\kc{\numConflictMatching}\kd contradiction visible only after a conversion or calculation\kc{\numConflictReasoning} \\[3pt]
    \tierrow{guideInk}{guideTint}{Guideline violations}{the page is correct but falls short of editorial or brand standards}
    \famname{guideInk}{Editorial}{coherence} & A sentence does not fit its passage. & Repeated meaning\kc{\numRepetitionPlanted}\kd wrong connective\kc{\numKindConnective}\kd slightly irrelevant sentence\kc{\numKindIrrelevant}\kd dangling reference\kc{\numKindDangling} \\
    \cmidrule{1-3}
    \famname{guideInk}{Formatting}{structure} & The markup a visitor sees is broken while the words are intact. & Heading shown as body text\kc{\numKindHeadingBody}\kd list item detached from its list\kc{\numKindListItem}\kd list flattened into a paragraph\kc{\numKindListFlat}\kd table cell missing\kc{\numKindCell}\kd heading demoted a level\kc{\numDemotedPlanted} \\
    \cmidrule{1-3}
    \famname{guideInk}{Brand voice}{voice} & A passage is written in a voice unlike the rest of its page, with every fact intact. & AI-sounding paraphrase\kc{\numKindSlopParaphrase}\kd emoji hype\kc{\numEmojiPlanted}\kd salesy urgency\kc{\numKindUrgency}\kd condescending or sarcastic tone\kc{\numKindAttitude}\kd wrong audience\kc{\numKindAudience}\kd AI scaffolding around the original wording\kc{\numKindSlopScaffolding}\kd stiff legalese\kc{\numKindLegalese} \\[2pt]
    \arrayrulecolor{guideInk}\bottomrule[1pt]
  \end{tabular}
  \arrayrulecolor{black}
  \caption{The five families of introduced issues, grouped into two tiers: \textcolor{errorInk}{\textbf{errors}}, where the page states something incorrect, and \textcolor{guideInk}{\textbf{guideline violations}}, where the page is correct but falls short of editorial or brand standards. Numbers in parentheses give how many issues of each kind were introduced in the scored test set of \numPlanted{} issues.}
  \label{tab:taxonomy}
\end{table*}

\section{\marketbench{}: Methodology}
\label{sec:benchmark}

\marketbench{} casts website auditing as a verifiable task with a known answer: the agent must detect each issue and correct it.
We take a snapshot of a real brand's website, introduce a set of issues into it whose locations and corrections we record, and ask an agent to audit the modified site.
Because every issue was introduced by us, each report the agent makes can be checked against ground truth.
\Cref{fig:hero} shows the pipeline; we formalize it below.

\paragraph{Formulation.}
A brand's website is a set of pages $\mathcal{W}=\{p_1,\dots,p_n\}$.
Each page is a sequence of blocks $p=(b_1,\dots,b_m)$, where a block is a paragraph, heading, list item or table row together with its markup, and every block has a stable anchor so that the agent, the answer key and the grader share one way to name a location.
We call the snapshot $\mathcal{W}$ the \emph{reference site} and treat its content as correct.
An \emph{issue} is an edit $e=(b, b', \phi)$ that replaces the content of one block $b$ with $b'$ and belongs to a family $\phi\in\Phi$ (\Cref{tab:taxonomy}).
A \emph{test site} is $\mathcal{W}'=\mathcal{W}\oplus\mathcal{E}$, the reference site with a set of issues $\mathcal{E}=\{e_1,\dots,e_K\}$ applied, at most two issues of a family per page and no two issues sharing a block.
Each issue carries an \emph{answer-key entry} $a_e=(L_e, d_e, F_e, s_e)$: the set of locations $L_e$ that identify it (two blocks for a cross-page conflict), a diagnosis $d_e$, the set of acceptable fixes $F_e$ and, for a wrong fact, an outside source $s_e$ that establishes the correct value.
The agent sees $\mathcal{W}'$ and a tool set $\mathcal{T}$ and returns a report $\mathcal{R}=\{r_1,\dots,r_M\}$, where each finding $r=(\ell, d, f, v)$ names a location $\ell$, a diagnosis $d$, a recommended fix $f$ and the evidence $v$ it rests on.
A grader $G$ scores $\mathcal{R}$ against $\{a_e\}_{e\in\mathcal{E}}$ as defined under \emph{Evaluation} below.
The answer key is never visible to the agent.

\paragraph{Site snapshots.}
Each of the \numBrands{} brands contributes \numPagesPerBrand{} pages from one surface of its site, such as a help center, product documentation or plan and pricing pages (\Cref{app:brands}).
A snapshot preserves each page's styles and images and removes anything that could take a live action, so the page the agent reads renders like the page a visitor sees.
Only what a visitor sees is in scope; capture details and exclusions are in \Cref{app:brands}.

\paragraph{Issue families.}
Issues fall into two tiers (\Cref{tab:taxonomy}).
\emph{Errors} make a page incorrect: a \emph{factual} error changes a claim that the outside world can settle, and a \emph{cross-page conflict} changes a claim so that it contradicts another page about the same entity, date, region and audience.
\emph{Guideline violations} leave a page correct but worse: an \emph{editorial} issue breaks how a sentence fits its passage, a \emph{formatting} issue breaks the markup a visitor sees while the words survive, and a \emph{brand voice} issue rewrites a passage in a voice unlike the rest of the page while keeping every fact.
The two tiers test different abilities.
Errors test whether an agent can establish what is true, from an outside source or from the rest of the site; guideline violations test editorial judgment, since nothing outside the page is needed to see them.
We introduce up to \numPerFamilyTarget{} issues of each family per site.

\paragraph{Introducing issues.}
Each issue is proposed by a generator model (GPT-6 Sol) as an exact edit to one block of the reference site, together with its diagnosis, its acceptable fixes and, for a factual error, the outside source that proves the original value.
The prompt asks for issues that require careful reading or one step of reasoning across the site yet are plainly wrong once pointed out, and it avoids famous facts, large rewrites and placeholders.
A screening model with web search checks every proposal and keeps it only when the edit introduces the diagnosed issue without repairing a pre-existing issue; it rejects conflicts whose two sides differ in scope and voice edits that change a fact.
Two constraints guard against ambiguity.
A factual error may only be introduced on a claim that an outside source settles, such as a law, a standard or a date in history, and the screen accepts it only when a quote from outside the brand's own domains proves both the correct claim and the error.
The brand's own claims, such as its prices, carry only cross-page conflicts, since no outside source can settle them (\Cref{app:planting}).

\paragraph{Human annotation.}
Expert annotators reviewed every proposed issue independently before it entered the answer key.
They agreed on \numRawAgreement\% of the proposed issues (Gwet's AC1 \numACOne{}; \citealp{gwet2008ac1}), and a third-party annotator settled each disagreement.
We dropped the proposals judged not to be issues, leaving \numPlanted{} issues across \numBrands{} sites.
We keep this test set private and plan to release a separate public set.

\paragraph{The audit task.}
The agent receives the task, the audit date, an index of the pages in $\mathcal{W}'$ and the tools $\mathcal{T}$: it can list and read pages, open a block's HTML, search the site, search the web and fetch outside pages, and it records findings one at a time (\Cref{app:budget}).
The audit date is the day we captured the site.
It is the date for which the agent and the grader judge every claim, so outside evidence counts only when it shows what held on that day.
This keeps the answer key from drifting as facts on the live web change after capture.
Every route to the live site is blocked, including the brand's domains, web archives and any page that reproduces the snapshot, so the agent cannot recover the original by looking it up (\Cref{app:grading}).
Each audit has a fixed budget of turns, tool calls, web searches and wall-clock time.

\paragraph{Evaluation.}
A grader built on GPT-6 Sol, run with one fixed prompt, first splits the report into atomic findings, one root cause each, then matches each finding to at most one issue.
A finding $r$ \emph{matches} an issue $e$ when its location falls in $L_e$ and its diagnosis agrees with $d_e$; a cross-page conflict must be anchored on both of its passages.
For every issue $e\in\mathcal{E}$ we record one of four outcomes:
\begin{itemize}[nosep,leftmargin=*]
  \item \textbf{Detected and corrected}: some finding matches $e$ and its fix $f$ passes two criteria, namely it is \emph{correct} (it lies in $F_e$ and, for a wrong fact, states the corrected value) and \emph{evidenced} (the finding's own evidence $v$ supports it, quoted from a page of $\mathcal{W}'$ or from a source the agent fetched).
  \item \textbf{Detected, weak correction}: some finding matches $e$ but its fix fails a criterion. Deleting the claim or asking someone to verify it counts here.
  \item \textbf{Misdiagnosed}: a finding points at a location in $L_e$ with the wrong diagnosis.
  \item \textbf{Not detected}: no finding points at $e$.
\end{itemize}
The grader also flags a finding as \emph{harmful} when its recommendation would make the page worse, including recommendations about content that was correct, and it treats the text of every finding as untrusted data, so an adversarial finding that addresses the grader earns nothing.
Findings that match no introduced issue do not count toward the score, though the grader can evaluate and accept them as real issues on the live sites.
Writing $c(e)\in\{0,1\}$ for whether $e$ was detected and corrected, a model's score on one site is
\begin{equation}
  S(\mathcal{W}') = \frac{1}{|\mathcal{E}|}\sum_{e\in\mathcal{E}} c(e),
\end{equation}
averaged over $k=3$ independent audits of that site, and the headline score averages $S$ over the \numBrands{} sites.
Because LLM graders can be biased \citep{wang2024fair,zheng2023judging}, human annotators audited a sample of the grader's decisions: they confirmed every issue it marked as not detected and \numHumanDetectAgree\% of the detections it credited, disagreeing mainly on fix quality, where the grader errs on the strict side (\Cref{app:human-review}).

\section{Experimental Setup}
\label{sec:setup}

\paragraph{Models.}
We evaluate seven frontier models from five labs: Gemini 4 Argon from Google \citep{google2026geminiargon}, Claude Opus 5.5 and Claude Fable 5.1 from Anthropic \citep{anthropic2026opus55,anthropic2026fable51}, Grok 4.7 from SpaceXSI \citep{xai2026grok47}, GPT-6 Sol and GPT-6 Astra from OpenAI \citep{openai2026gpt6sol,openai2026gpt6astra} and Muse Wonder from Meta \citep{meta2026musewonder}.
To study how model size trades against cost, we also evaluate three smaller models, Claude Sonnet 5.5 \citep{anthropic2026sonnet55}, Claude Haiku 5.5 \citep{anthropic2026haiku55} and GPT-6 Luna \citep{openai2026gpt6sol}.
We run each model through its provider's API at the default reasoning effort.

\paragraph{Agent harness.}
To ensure that differences in score come from the model alone, we run every model in the same harness with the same eight tools (\Cref{app:budget}).
Four read the test site (\texttt{list\_pages}, \texttt{read\_page}, \texttt{read\_html} and \texttt{search\_site}), two reach the web (\texttt{web\_search} and \texttt{fetch\_external}), and two write the report: \texttt{record\_finding} records one finding with its block, diagnosis and fix, and \texttt{submit\_audit} ends the audit.
All models search through the same \texttt{web\_search} tool, thus removing differences between the search tools that providers build into their APIs.
The agent has no shell, no file system and no access to the answer key.
The agent is provided the task prompt, the audit date, the entry URL and the index of the test-site pages in its context.
Each audit may use at most \numMaxTurns{} turns, \numMaxToolCalls{} inspection tool calls, \numMaxSearches{} web searches and \numMaxMinutes{} minutes, and an audit that reaches any of these limits ends there.
The agent is informed of the budget it has left after every turn.

\paragraph{Blocking leakage.}
\label{sec:blocks}
Every test site is built from public pages, so the unedited original of each page is still online, on the brand's own site and in web archives.
An agent that finds the original can spot each introduced issue by comparing the two pages, without judging the content at all.
Such leaks are known to inflate benchmark scores, and search agents can find them at test time \citep{sainz2023trouble,han2025searchtime}.
We therefore block every route to the original pages.
Both web tools refuse the brand's domains (\Cref{sec:benchmark}), \numWebArchives{} web archives and caches, and any page that reproduces a reference page (\Cref{app:grading}).
Across its three runs, each frontier model made between \numCheatCallsMin{} and \numCheatCallsMax{} tool calls that tried to reach the live site or a copy of it, and the harness blocked \numCheatBlockedPct\% of these calls.
The \numCheatCallsThrough{} that got through were site searches by Sol and Grok whose result snippets quoted a copy of a page, too few to change either model's rank.

\paragraph{Runs and scoring.}
An agent's score varies from run to run, so we run every model on every brand three times ($k=3$) with the same prompt, tools and budget.
We grade every audit with the grader and outcomes defined in \Cref{sec:benchmark}, and report the share of introduced issues that are detected and corrected, averaged over the three runs and then over brands.

\section{Results}
\label{sec:results}

\subsection{Frontier models fail to detect all issues}
\label{sec:headline}

\begin{figure}[!htb]
  \centering
  \includegraphics[width=\linewidth]{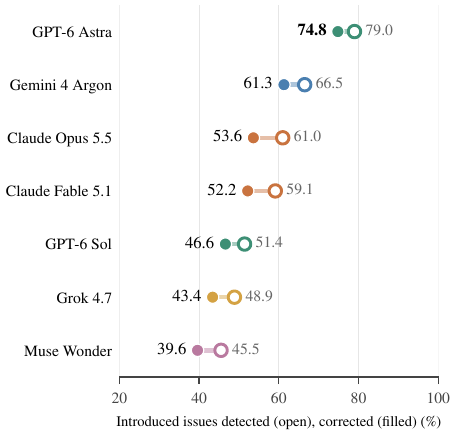}
  \caption{Astra detects and corrects the most introduced issues. Share of the \numPlanted{} introduced issues each model detected (open dot) and corrected (filled dot), averaged over runs and brands.}
  \label{fig:headline}
\end{figure}

\Cref{fig:headline} and \Cref{tab:scores} give each model's detected and corrected shares.
The best model, Astra, detects \numFoundAstra\% of the introduced issues and corrects \numScoreAstra\%, while the second-best model, Argon, detects \numFoundArgon\% and corrects \numScoreArgon\%.
Interestingly, Astra corrects the most issues in every family but formatting, where Argon corrects \numFamFixedFormattingArgon\% against Astra's \numFamFixedFormattingAstra\%.
Even the best model, Astra, leaves \numBestUnfixed\% of the introduced issues uncorrected, thus leaving substantial headroom on \marketbench{}.
Agents also reported about \numRealErrors{} real issues on the live sites that we did not introduce, which the grader accepted (\Cref{app:real-errors}).

\begin{table*}[t]
  \centering
  \small
  \begin{tabular}{@{}lr@{\hspace{7pt}}!{\color{ruleGray}\vrule}@{\hspace{7pt}}rr@{\hspace{14pt}}rrr@{\hspace{14pt}}r@{}}
    \toprule
    & & \multicolumn{2}{c}{Errors} & \multicolumn{3}{c}{Guideline violations} & \\
    \cmidrule(lr){3-4}\cmidrule(lr){5-7}
    Model & Total detected & Factual & Cross-page conflict & Editorial & Formatting & Brand voice & Total corrected \\
    \midrule
    Astra & \textbf{\numFoundAstra} & \textbf{\numFamFixedFactualAstra} & \textbf{\numFamFixedConflictAstra} & \textbf{\numFamFixedEditorialAstra} & \numFamFixedFormattingAstra & \textbf{\numFamFixedVoiceAstra} & \textbf{\numScoreAstra} \\
    Argon & \numFoundArgon & \numFamFixedFactualArgon & \numFamFixedConflictArgon & \numFamFixedEditorialArgon & \textbf{\numFamFixedFormattingArgon} & \numFamFixedVoiceArgon & \numScoreArgon \\
    Opus & \numFoundOpus & \numFamFixedFactualOpus & \numFamFixedConflictOpus & \numFamFixedEditorialOpus & \numFamFixedFormattingOpus & \numFamFixedVoiceOpus & \numScoreOpus \\
    Fable & \numFoundFable & \numFamFixedFactualFable & \numFamFixedConflictFable & \numFamFixedEditorialFable & \numFamFixedFormattingFable & \numFamFixedVoiceFable & \numScoreFable \\
    Sol & \numFoundSol & \numFamFixedFactualSol & \numFamFixedConflictSol & \numFamFixedEditorialSol & \numFamFixedFormattingSol & \numFamFixedVoiceSol & \numScoreSol \\
    Grok & \numFoundGrok & \numFamFixedFactualGrok & \numFamFixedConflictGrok & \numFamFixedEditorialGrok & \numFamFixedFormattingGrok & \numFamFixedVoiceGrok & \numScoreGrok \\
    Wonder & \numFoundWonder & \numFamFixedFactualWonder & \numFamFixedConflictWonder & \numFamFixedEditorialWonder & \numFamFixedFormattingWonder & \numFamFixedVoiceWonder & \numScoreWonder \\
    \bottomrule
  \end{tabular}
  \caption{Astra corrects the most issues in every family except formatting, where Argon leads. Share of introduced issues each model detected and corrected, averaged over runs and brands. The family columns give the share of that family's issues corrected. Bold marks the best model in each column.}
  \label{tab:scores}
\end{table*}

\paragraph{Formatting detection is concentrated in two models.}
We find that formatting issues are best corrected by Argon (\numFamFixedFormattingArgon\%) and Astra (\numFamFixedFormattingAstra\%) (\Cref{tab:scores}).
Sol (\numFamFixedFormattingSol\%) and Fable (\numFamFixedFormattingFable\%) are on par, and Opus corrects only \numFamFixedFormattingOpus\% of formatting issues.

We dive deeper into model audit habits, read from the traces of all runs and summarized in \Cref{app:extra}.

\subsection{Models miss issues in text they read}
\label{sec:funnel}

\begin{figure}[t]
  \centering
  \includegraphics[width=\linewidth]{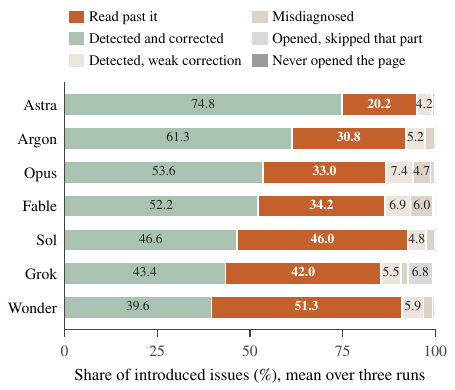}
  \caption{Agents read the text behind most of the issues they leave undetected. Each bar splits a model's \numPlanted{} introduced issues by how far the agent got with each, as a share averaged over runs and brands. ``Read past it'' means the agent read the block but never detected the issue.}
  \label{fig:funnel}
\end{figure}

\Cref{fig:funnel} splits each model's introduced issues by how far the agent got with them, from never opening the page to detecting and correcting the issue.
Here, we say an agent \emph{reads past} an introduced issue when it reads the block that holds the issue but fails to detect it.
Reaching the text is rarely the problem.
Of the models evaluated, Grok leaves the most introduced issues unread, at \numNotReadGrokPct\%.
The rest leave at most \numNotReadMaxPct\% unread.
The share of introduced issues that agents read past is larger, between \numReadPastMin\% (Astra) and \numReadPastMax\% (Wonder).

Across families, the share of issues each model reads past (\Cref{tab:readpast-family}) shows that models most often fail to detect guideline violations.
Editorial (\numFamReadPastEditorialAll\%) and formatting issues (\numFamReadPastFormattingAll\%) are read past the most, on average across the models.

\begin{table}[!htb]
  \centering
  \small
  \setlength{\tabcolsep}{4pt}
  \begin{tabular}{@{}lccccc@{}}
    \toprule
     & \multicolumn{2}{c}{Errors} & \multicolumn{3}{c}{Guideline violations} \\
    \cmidrule(lr){2-3}\cmidrule(l){4-6}
    Model & Factual & Conflict & Editorial & Format & Voice \\
    \midrule
    Astra & \textbf{\numFamReadPastFactualAstra} & \numFamReadPastConflictAstra & \textbf{\numFamReadPastEditorialAstra} & \numFamReadPastFormattingAstra & \textbf{\numFamReadPastVoiceAstra} \\
    Argon & \numFamReadPastFactualArgon & \numFamReadPastConflictArgon & \numFamReadPastEditorialArgon & \textbf{\numFamReadPastFormattingArgon} & \numFamReadPastVoiceArgon \\
    Opus & \numFamReadPastFactualOpus & \textbf{\numFamReadPastConflictOpus} & \numFamReadPastEditorialOpus & \numFamReadPastFormattingOpus & \numFamReadPastVoiceOpus \\
    Fable & \numFamReadPastFactualFable & \numFamReadPastConflictFable & \numFamReadPastEditorialFable & \numFamReadPastFormattingFable & \numFamReadPastVoiceFable \\
    Sol & \numFamReadPastFactualSol & \numFamReadPastConflictSol & \numFamReadPastEditorialSol & \numFamReadPastFormattingSol & \numFamReadPastVoiceSol \\
    Grok & \numFamReadPastFactualGrok & \numFamReadPastConflictGrok & \numFamReadPastEditorialGrok & \numFamReadPastFormattingGrok & \numFamReadPastVoiceGrok \\
    Wonder & \numFamReadPastFactualWonder & \numFamReadPastConflictWonder & \numFamReadPastEditorialWonder & \numFamReadPastFormattingWonder & \numFamReadPastVoiceWonder \\
    \midrule
    All models & \numFamReadPastFactualAll & \numFamReadPastConflictAll & \numFamReadPastEditorialAll & \numFamReadPastFormattingAll & \numFamReadPastVoiceAll \\
    \bottomrule
  \end{tabular}
  \caption{Models read past editorial and formatting issues most often and cross-page conflicts least. Share of each family's introduced issues that each model read past (\%), averaged over runs and brands. Lower is better, since the agent detected more of the issues whose text it read. Bold marks the lowest in each family.}
  \label{tab:readpast-family}
\end{table}

\subsection{Model size and cost}
\label{sec:price}

\begin{figure}[!h]
  \centering
  \includegraphics[width=\linewidth]{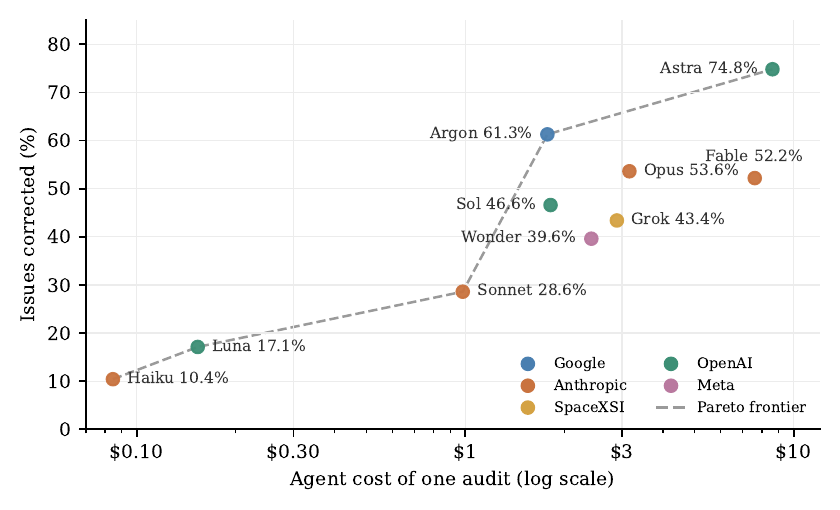}
  \caption{Share of introduced issues corrected against the agent's average cost of one audit (log scale). The dashed line joins the models on the Pareto frontier, where no other model scores higher for less cost per audit. Astra scores highest at \$\numAuditCostMax{} per audit, while Argon reaches \numScoreArgon\% at \$\numAuditCostMin{}, about a fifth of Astra's cost.}
  \label{fig:pareto}
\end{figure}

When we plot each model's score against the agent's cost of one audit (\Cref{fig:pareto}; exact values in \Cref{tab:cost}, \Cref{app:budget}), Astra leads the Pareto frontier with the highest score.
Smaller models cost far less but correct far fewer issues.
The second-best model on the Pareto frontier, Argon, costs \$\numAuditCostMin{} per audit, about a fifth of Astra's \$\numAuditCostMax{}.
Fable scores close to Opus at more than twice its cost per audit.
For the three smaller models we tested, from two labs, performance does not hold up at lower cost.

\FloatBarrier
\subsection{Ablations}
\label{sec:ablations}

\subsubsection{Web search}
\label{sec:web}

We investigate the impact of web search tools on agent performance by running Opus, Sol and Argon with \texttt{web\_search} and \texttt{fetch\_external} removed (\Cref{tab:web}).
Our methodology (\Cref{sec:benchmark}) considers a factual error corrected only when its fix $f$ states the corrected value and the finding's own evidence $v$ supports it, quoted from a page of $\mathcal{W}'$ or from a source the agent fetched.
Without web access, models cannot fetch the outside source that a factual fix needs as evidence.
Hence, we score \emph{right replacement}, which checks whether the agent's proposed replacement matches the answer key without checking the evidence.
Without the web, the share of factual errors corrected drops by \numWebFactsLostPctMin{} to \numWebFactsLostPctMax{} points, while the right-replacement share drops by only \numWebRightDropArgon{} to \numWebRightDropSol{} points.
These drops are within the variation between runs, since a model's right-replacement share without the web varies between runs by up to \numWebRightRunGapOffMax{} points.
This suggests that for the factual errors they detect, the models already know the corrected fact, and the web gives them a source to cite as evidence.
Their searches corroborate this.
When a model corrected a factual error with the web on, it typed the corrected fact into a search before any result showed it \numSearchKnewPct\% of the time, and another \numSearchNoWebPct\% of these corrections needed no web call at all.
Fewer than 1\% came from a chain of searches that turned up the fact (\Cref{tab:hops}).
Citing that source is where Opus loses the most: with the web on, its right-replacement share exceeds its corrected share by \numWebRightGapOnOpus{} points, against \numWebRightGapOnSol{} for Sol and \numWebRightGapOnArgon{} for Argon.

\begin{table}[t]
  \centering
  \small
  \begin{tabular}{@{}lcccc@{}}
    \toprule
     & \multicolumn{4}{c}{Factual errors (\%)} \\
    \cmidrule(l){2-5}
     & \multicolumn{2}{c}{Corrected} & \multicolumn{2}{c}{Right replacement} \\
    \cmidrule(lr){2-3}\cmidrule(l){4-5}
    Model & web on & web off & web on & web off \\
    \midrule
    Opus & \numWebCorrectedOnOpus & \numWebCorrectedOffOpus{} ($-$\numWebCorrectedDropOpus) & \numWebRightOnOpus & \numWebRightOffOpus{} (\textbf{$-$\numWebRightDropOpus}) \\
    Sol & \numWebCorrectedOnSol & \numWebCorrectedOffSol{} ($-$\numWebCorrectedDropSol) & \numWebRightOnSol & \numWebRightOffSol{} (\textbf{$-$\numWebRightDropSol}) \\
    Argon & \numWebCorrectedOnArgon & \numWebCorrectedOffArgon{} ($-$\numWebCorrectedDropArgon) & \numWebRightOnArgon & \numWebRightOffArgon{} (\textbf{$-$\numWebRightDropArgon}) \\
    \bottomrule
  \end{tabular}
  \caption{Removing the web sharply cuts the corrected share but barely moves the right-replacement share. Corrected requires the finding's evidence to support the corrected value; right replacement checks only that the fix states the corrected value. Shares of the \numWebFactsPlanted{} factual errors, averaged over runs and brands. Bold right-replacement drops are within the variation between runs (up to \numWebRightRunGapOffMax{} points).}
  \label{tab:web}
\end{table}

\subsubsection{Reasoning effort}
\label{sec:effort}
We investigate the contribution of reasoning effort to performance by running Sol and Opus at medium (default), high and xhigh effort over all brands (\Cref{tab:effort}).
More reasoning effort \textit{marginally} raises both detection and correction, by up to \numEffortDetGainSol{} and \numEffortCorrGainSol{} points for Sol at xhigh and \numEffortDetGainOpus{} and \numEffortCorrGainOpus{} points for Opus at high, but Opus falls back at xhigh.
We also double the turn budget and other limits for Sol and Opus; \Cref{app:turns} documents the differences.
\begin{table}[t]
  \centering
  \small
  \setlength{\tabcolsep}{5pt}
  \begin{tabular}{@{}llrr@{}}
    \toprule
    Model & Effort & Detected (\%) & Corrected (\%) \\
    \midrule
    Sol & medium (default) & 51.4 & 46.6 \\
    Sol & high & 52.6 & 48.4 \\
    Sol & xhigh & \textbf{56.1} & \textbf{53.5} \\
    \midrule
    Opus & medium (default) & 61.0 & 53.6 \\
    Opus & high & \textbf{66.0} & \textbf{58.4} \\
    Opus & xhigh & 63.5 & 58.1 \\
    \bottomrule
  \end{tabular}
  \caption{Sol corrects the most at xhigh and Opus at high. Reasoning-effort ablation. Higher is better in both columns, and bold marks each model's best setting. Averaged over brands.}
  \label{tab:effort}
\end{table}

\FloatBarrier
\subsubsection{Grader}
\label{sec:grader-ablation}
In recent times, decision models, which classify among a fixed set of options instead of generating text, have been adopted to judge text at a fraction of the cost of a generative LLM judge \citep{li2026jev,openai2026decisions}.
Once a report is split into findings, our grader makes three decisions for each finding $r=(\ell,d,f,v)$ (\Cref{sec:benchmark}): whether $r$ matches an issue $e$, choosing one issue of $\mathcal{E}$ or none; whether its fix $f$ is correct for $e$; and whether its evidence $v$ supports $f$.
To test whether a decision model can be used for the grader, we let JEV take over these three decisions on all three runs (\Cref{tab:grader}).
JEV sees the answer key and each finding's page context, similar to our grader (\Cref{app:jev}).
We compare the two graders decision by decision over the findings from the runs (\Cref{tab:grader}).
JEV agrees with Sol on whether a finding matches an issue (Cohen's $\kappa$ \numJevMatchKappa).
However, JEV rarely rejects a fix: it passes 99.0\% of fixes as correct, against 92.6\% for Sol, making it less effective as a judge.
JEV's decisions cost \$0.017 per audit, about 36 times cheaper than the \$\numGradingCost{} for the Sol grader, so a decision model that can judge fixes would make grading far cheaper (\Cref{sec:futurework}).
\begin{table}[t]
  \centering
  \small
  \begin{tabular}{@{}lrrr@{}}
    \toprule
    Decision & \makecell[r]{Sol yes\\(\%)} & \makecell[r]{JEV yes\\(\%)} & $\kappa$ \\
    \midrule
    Matches an issue & 67.8 & 71.7 & 0.89 \\
    Fix is correct & 92.6 & 99.0 & \textbf{0.16} \\
    Fix is evidenced & 94.8 & 92.1 & \textbf{0.16} \\
    \bottomrule
  \end{tabular}
  \caption{Decisions of the Sol grader and JEV on the same findings from all three runs of the seven frontier models. Correctness and evidence are compared on the findings both graders match to the same issue. JEV matches findings as Sol does but passes almost every fix, so it rarely catches the incorrect ones. Bold marks the low agreement on the fix decisions.}
  \label{tab:grader}
\end{table}

\FloatBarrier

\section{Related Work}
\label{sec:related-work}

\paragraph{Marketing evaluations.}
Prior work evaluates language models on individual marketing skills: marketing knowledge in textbook questions \citep{burke2026ami}, ad text \citep{zhang2025adtec,mita2024camera}, advertising analytics \citep{hu2026adbench}, ads inserted into chatbot answers \citep{hu2026gembench}, creative ideas judged by human preference \citep{bhat2025creativity} and influencer search \citep{chen2025xbench}.
Each of these benchmarks targets one skill in isolation, and none tests whether a model can keep a brand's published content correct, consistent and on brand.
\marketbench{} evaluates this content directly, on real brand websites that frontier models now both write and read, and keeps every answer verifiable by introducing known issues into the pages.

\paragraph{Agent benchmarks.}
Agent benchmarks measure whether an agent completes tasks on real or self-hosted websites \citep{deng2023mind2web,zhou2024webarena}, answers questions that need browsing \citep{mialon2024gaia,wei2025browsecomp}, serves a user under a domain policy \citep{yao2025taubench}, or resists prompt injections hidden in tool outputs \citep{debenedetti2024agentdojo}.
They treat the pages the agent reads as ground truth, while \marketbench{} asks the agent to find honest errors in them.

\paragraph{Detecting errors in text.}
Fact-checking and conflicting-context benchmarks hand the model a claim or a pair of passages \citep{thorne2018fever,schlichtkrull2023averitec,wu2024clasheval,hou2024wikicontradict,vu2024freshllms}, error-detection benchmarks place errors in a single document \citep{li2024contradoc,lovering2026find,xi2025flaws,son2025spot,bianchi2025toerr}, and editing benchmarks name the edit to make \citep{raheja2023coedit,dwivediyu2022editeval}.
\marketbench{} spreads issues across a whole site, includes conflicts between pages and facts that only outside sources settle, and grades the fix along with the finding.

\paragraph{Denoising and error correction.}
Denoising objectives corrupt an input and train a model to recover it, from denoising autoencoders \citep{vincent2008denoising} to masked language modeling \citep{devlin2019bert}, replaced-token detection \citep{clark2020electra} and sequence-to-sequence denoising \citep{lewis2020bart,raffel2020t5}.
Grammatical error correction applies the same detect-then-correct structure to text written by people \citep{bryant2023gec}.
\marketbench{} uses this structure for evaluation: its corruptions are semantic edits written to read like real content, spread across a whole site, and each comes with a known correction.

\section{Conclusion}
\label{sec:conclusion}
We introduce \marketbench{}, the first open-source benchmark of the marketing ability of frontier models.
Evaluating marketing ability is challenging, as the quality of a marketing output depends on brand and taste.
To address this, we frame the evaluation as a denoising task in which an agent must detect and correct issues introduced into real brand websites, and since the clean version of every page is known, both steps admit a definite answer.
We find that frontier models are far from solving \marketbench{}: the best model, GPT-6 Astra, detects and corrects \numScoreAstra\% of the introduced issues, and most issues that go undetected sit in text the model has already read.
We hope \marketbench{} enables the community to faithfully measure and improve the marketing ability of frontier models, and we plan to release a separate public set to support this, as a first step towards frontier marketing models.

\section{Future Work}
\label{sec:futurework}

\paragraph{Grading with decision models.}
Grading every audit with a generative model keeps the benchmark costly to run often, and too costly to serve as an environment for reinforcement learning.
Advances in decision models that improve their accuracy and their context windows would make them viable for grading and cut the cost of running the benchmark (\Cref{sec:grader-ablation}).

\paragraph{Closing the read-past gap.}
Most issues that agents miss sit in text they already read (\Cref{sec:funnel}).
Methods that make an agent check each passage it reads, through training or through the harness, are a direct way to raise detection.

\paragraph{Wider coverage.}
\marketbench{} covers \numBrands{} English brand websites captured at one point in time.
Extending it to more brands, languages, regions and site types, and to live sites over time, would test how these results generalize.

\bibliography{refs}
\bibliographystyle{acl_natbib}

\appendix
\section{Brands and cached sites}
\label{app:brands}

The test set covers \numBrands{} brands across retail, travel, software, finance and streaming (\Cref{tab:brands}).
Each brand's pages come from one evaluated surface, such as help centers, policies and terms, product documentation, plan and pricing pages, or newsrooms.
Only what a visitor sees on the rendered page is in scope, so metadata such as SEO descriptions and issues visible only in the HTML are out.

\begin{table}[H]
  \centering
  \small
  \begin{tabular}{@{}>{\raggedright\arraybackslash}p{0.32\linewidth}>{\raggedright\arraybackslash}p{0.62\linewidth}@{}}
    \toprule
    Brand & Evaluated surface \\
    \midrule
    Airbnb & Help-center policy articles \\
    American Airlines & US customer-service and travel-information pages \\
    Apple & iCloud and Apple Account support articles \\
    Atlassian & Confluence Cloud support documentation \\
    Canva & Product marketing, policies and newsroom \\
    IKEA (US) & US customer-service policies, terms and help \\
    Microsoft Outlook & Outlook.com support articles \\
    Nike & Corporate pages, help articles and membership \\
    Oracle & OCI Object Storage documentation \\
    Ramp & Pricing, product, release and announcement pages \\
    Spotify (US) & US plan pages, US legal pages and newsroom \\
    \bottomrule
  \end{tabular}
  \caption{The \numBrands{} test-set brands and the part of each site we cached, \numPagesPerBrand{} pages each.}
  \label{tab:brands}
\end{table}

\paragraph{Page selection.}
Each brand contributes \numPagesPerBrand{} pages, chosen from a list of deep seed URLs on the brand's own domains and the in-scope pages they link to.

\paragraph{Capture.}
We captured every site from a US network on September 24, 2026, so region-specific pages and prices match what a US visitor sees, and we judge staleness against that date.
The capture fetches server-rendered HTML, honors each site's \texttt{robots.txt}, and rejects error pages, bot walls and pages with too little text.
It keeps the page's styles and images, removes scripts, disables forms and rewrites links, so the cached page renders like the live page and takes no live action.

\paragraph{Blocks.}
The capture splits every page into anchored paragraph blocks, which give the agent, the answer key and the grader one shared way to name a location.
The agent reads headings, list items and table rows through text markers, and it can read any block's HTML.

\section{Introducing issues and answer keys}
\label{app:planting}

\paragraph{Kinds within each family.}
\begin{itemize}
  \item \textbf{Cross-page conflict} changes one passage so it contradicts another about the same entity, date, region and audience.
  \item \textbf{Editorial} adds a sentence that repeats another in different words, a connective that reverses the logic, a pointer to a table or section the page lacks, or an on-brand sentence that does not belong in its passage.
  \item \textbf{Formatting} breaks markup a visitor sees while the words survive, such as a heading shown as body text, a list run into one paragraph or a missing table cell.
\end{itemize}

\paragraph{Generation.}
The generator, GPT-6 Sol, proposes each introduced issue as an exact edit to one anchored block of a cached page, with its diagnosis, its acceptable fixes and, for a factual error, the outside source that proves the original value.
It aims for hard issues, so it avoids famous facts, large changes and placeholders, and it puts at most two issues of one family on a page.

\paragraph{Screening.}
An automated screen, also GPT-6 Sol with web search, keeps an edit only when it introduces the diagnosed issue without repairing a pre-existing issue, and it rejects conflicts whose sides differ in scope and voice edits that change a fact.
No two introduced issues share a location, and one wrong fact introduced in several places is a single answer-key entry where naming any place counts.

\paragraph{Outside sources.}
A factual error may only sit on a claim the outside world settles, such as a law, a standard or a date in history, because the live site would show the correct value.
The screen accepts it only when a quote from outside the brand's domains proves both the correct claim and the error, and all \numFamFactual{} scored factual errors carry such a source from \numFactSourceHosts{} distinct hosts, archived so the grader checks fixes against the same text.
That leaves \numPlanted{} answer-key entries.

\section{Harness, budget and costs}
\label{app:budget}

\paragraph{Tools.}
The agent works through these tools:
\begin{itemize}
  \item \texttt{list\_pages} lists every test-site page in scope.
  \item \texttt{read\_page} reads a page's text with its paragraph anchors, in chunks the agent pages through.
  \item \texttt{read\_html} shows the test-site HTML of one block and up to three neighbors on each side, for checking markup and writing markup fixes.
  \item \texttt{search\_site} searches the test site only.
  \item \texttt{web\_search} searches the web and returns the top \numSearchResults{} results, each with a short excerpt. It is backed by Parallel's Search API \citep{parallelsearch}, since not all providers' own search tools can block domains.
  \item \texttt{fetch\_external} fetches and archives an outside page, which the agent can then cite.
  \item \texttt{record\_finding} records one finding as soon as it is ready, and the recorded findings are the report. A finding names a category, its locations (page, paragraph anchor and quote), a diagnosis, evidence, a priority, a confidence, one recommended action and, optionally, the whole paragraph as it should read. The prompt asks for one finding per root cause.
  \item \texttt{submit\_audit} ends the audit with a summary and may withdraw recorded findings.
\end{itemize}
Long audits outgrow a model's context, so older history is compacted.
The agent saves structured checkpoints and rereads archived sources with \texttt{read\_evidence}, and its recorded findings are always kept.
An audit that ends without \texttt{submit\_audit} is scored on the findings it recorded.

\paragraph{Budget.}
Beyond the limits in \Cref{sec:setup}, each audit may use at most \numMaxInputTokens{} input tokens and \numMaxOutputTokens{} output tokens in all, \numResponseTokens{} output tokens per response and a context of \numContextTokens{} tokens.
Recording and submitting findings do not count against the inspection calls.
On its last turn only \texttt{record\_finding} and \texttt{submit\_audit} are offered, so every audit can record its findings before the budget runs out.
Each model call may take \numCallSeconds{} seconds and gets \numCallAttempts{} attempts.
Timeouts, rate limits and server errors are retried, and time spent waiting out a rate limit or server error is given back to the audit.

\paragraph{Costs.}
Every model call goes through one spend ledger priced from the makers' published rates, and each web search costs \$\numSearchPrice{}.
An audit costs the agent \$\numAuditCostMin{} to \$\numAuditCostMax{} for the seven frontier models and \$\numAuditCostHaiku{} to \$\numAuditCostSonnet{} for the smaller ones (\Cref{tab:cost}).
Grading adds about \$\numGradingCost{} per audit, with the answer key sent as its own cached message so most of the grader's input is read from cache.

\begin{table}[H]
  \centering
  \small
  \begin{tabular}{@{}lrr@{}}
    \toprule
    Model & Corrected (\%) & Agent cost per audit (\$) \\
    \midrule
    Astra & \textbf{\numScoreAstra} & \textbf{\numAuditCostMax} \\
    Argon & \textbf{\numScoreArgon} & \textbf{\numAuditCostMin} \\
    Opus & \numScoreOpus & \numAuditCostOpus \\
    Fable & \numScoreFable & \numAuditCostFable \\
    Sol & \numScoreSol & \numAuditCostSol \\
    Grok & \numScoreGrok & \numAuditCostGrok \\
    Wonder & \numScoreWonder & \numAuditCostWonder \\
    \midrule
    Sonnet & \textbf{\numScoreSonnet} & \textbf{\numAuditCostSonnet} \\
    Luna & \textbf{\numScoreLuna} & \textbf{\numAuditCostLuna} \\
    Haiku & \textbf{\numScoreHaiku} & \textbf{\numAuditCostHaiku} \\
    \bottomrule
  \end{tabular}
  \caption{Share of introduced issues each model corrected and the agent's cost of one audit, averaged over runs and brands. Bold marks the models on the Pareto frontier (\Cref{fig:pareto}).}
  \label{tab:cost}
\end{table}

\section{Grader inputs and blocking}
\label{app:grading}

\paragraph{Grader inputs.}
The grader sees the test-site pages, the answer-key entries and the archived sources, and a deterministic scorer turns its labels into scores.

\paragraph{Blocking.}
\texttt{fetch\_external} refuses the brand's domains, including its owned domains elsewhere, and \numWebArchives{} web archives and caches, and it refuses an outside URL that redirects to any of them.
It also refuses any fetched page that reproduces \numCopyPassages{} or more passages of at least \numCopyChars{} characters from one cached page, which catches mirrors and proxies on other hosts.
\texttt{web\_search} excludes the same domains and archives at the search provider, and the harness drops any that remain before the model sees the results.

\paragraph{JEV grader.}
\label{app:jev}
JEV (\texttt{jev-1.13.0}) receives the audit date and the answer-key entries $\{a_e\}_{e\in\mathcal{E}}$, without the sources $s_e$ or the page excerpts around $L_e$, since JEV accepts at most 32K tokens per decision.
For each finding $r=(\ell,d,f,v)$ it also receives the page context the Sol grader saw for $r$.
Around 2.2\% of decisions are made without this context, as they exceed JEV's context limit.
JEV returns a probability for each option, and we take the most likely issue as the match and count a fix as correct or evidenced when its probability is at least 0.5.
It makes the three decisions as follows.
\begin{itemize}
  \item \textbf{Matches an issue.} Which $e\in\mathcal{E}$, if any, does $r$ identify? It matches $e$ when $\ell\in L_e$ and $d$ describes the problem in $d_e$ in the same direction, with both passages for a cross-page conflict.
  \item \textbf{Fix is correct.} For a matched $e$, does $f$ leave the page fixed? It must agree with a fix in $F_e$, keep every fact $e$ does not concern and, for a factual error, state the corrected value.
  \item \textbf{Fix is evidenced.} For a matched $e$, does $v$ support $f$? For a factual error $v$ must support the corrected value, while a fix that adds no new fact is supported by its quoted locations.
\end{itemize}

\section{Human review of the grader}
\label{app:human-review}

Three annotators reviewed 60 of the grader's decisions from the last run, 10 from each of six classes, without seeing the model or the grader's verdict (\Cref{tab:human-review}).
They confirmed every issue the grader marked as missed and \numHumanDetectAgree\% of the detections it credited.
Most disagreements were about fix quality, where the grader errs on the strict side, and weak corrections are only \numWeakShare\% of credited findings, so the ranking of the models would not change even if every one counted as corrected (\Cref{tab:scores}).
For unplanted findings, annotators agreed with \numHumanAcceptedAgree\% of the real issues the grader accepted and \numHumanRejectedAgree\% of those it rejected, so our count of real issues on the live sites is conservative.

\begin{table}[H]
  \centering
  \small
  \begin{tabular}{@{}lc@{}}
    \toprule
    Grader decision & Annotator agreed (\%) \\
    \midrule
    Detected (corrected or weak) & \textbf{\numHumanDetectAgree} \\
    \midrule
    Corrected & \numHumanFixedAgree \\
    Weak correction & \numHumanWeakAgree \\
    Misdiagnosed & \numHumanWrongAgree \\
    Missed & \textbf{\numHumanMissedAgree} \\
    Real issue, accepted & \numHumanAcceptedAgree \\
    Real issue, rejected & \numHumanRejectedAgree \\
    \bottomrule
  \end{tabular}
  \caption{Share of each class of grader decisions the annotator agreed with, from 10 decisions per class (8 answered for rejected real issues). The first row pools the corrected and weak-correction decisions and asks only whether the issue was detected. Weak corrections are \numWeakShare\% of all the corrections the grader credited. Bold marks the detection decisions, which the annotators confirmed. The rows below them judge fix quality, where most disagreements fall.}
  \label{tab:human-review}
\end{table}

\section{Real issues on the live sites}
\label{app:real-errors}

While auditing the test sites, agents also reported issues that we did not introduce.
The grader accepted about \numRealErrors{} distinct ones, across \numRealErrorPages{} pages of all \numBrands{} brands.
Most are visible to any reader, such as typos, doubled words, numbers that disagree across pages and stale facts.
The examples below were present on the linked pages when we captured the sites on September 24, 2026.

\begin{figure}[H]
  \centering
  \includegraphics[width=\linewidth]{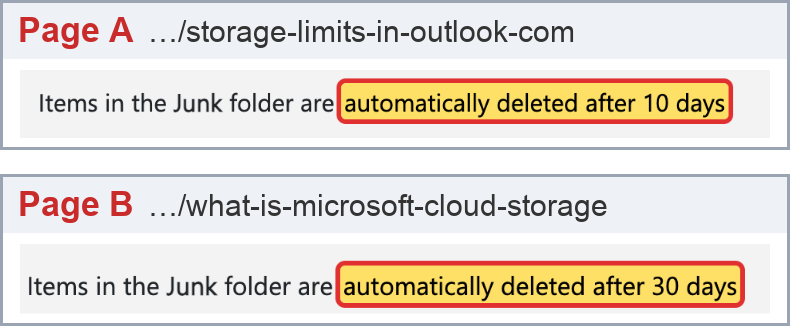}
  \caption{Two Microsoft support pages that disagree on how long Outlook keeps junk email, as captured on September 24, 2026.}
  \label{fig:outlook}
\end{figure}

\begin{itemize}[itemsep=2pt,topsep=4pt]
  \item \textbf{Outlook.} One support page says items in the Junk folder are deleted after 10 days, while the cloud storage page and one other say 30 days (\Cref{fig:outlook}; \href{https://support.microsoft.com/en-us/outlook/storage-limits-in-outlook-com}{storage limits}, \href{https://support.microsoft.com/en-us/onedrive/what-is-microsoft-cloud-storage}{cloud storage}, \href{https://support.microsoft.com/en-us/outlook/email-is-missing-from-your-outlook-com-inbox}{missing email}).
  \item \textbf{Oracle.} The page on changing a bucket's visibility swaps the definitions of \texttt{ObjectRead} and \texttt{ObjectReadWithoutList}, so it contradicts the bucket creation page (\href{https://docs.oracle.com/en-us/iaas/Content/Object/Tasks/managingbuckets_topic-To_change_the_visibility_of_a_bucket.htm}{visibility}, \href{https://docs.oracle.com/en-us/iaas/Content/Object/Tasks/managingbuckets_topic-To_create_a_bucket.htm}{bucket creation}).
  \item \textbf{Canva.} The Terms of Use still carry a legal drafting placeholder in square brackets in the section on Cavalry software (\href{https://www.canva.com/policies/terms-of-use/}{terms of use}).
  \item \textbf{Ramp.} A blog post gives a 4.13\% yield in one place and 4.63\% in another, and a promotional sentence is spliced into the middle of a link, splitting ``support page'' into ``suppo'' and ``rt page'' (\href{https://ramp.com/blog/introducing-the-new-investment-account}{investment account post}).
  \item \textbf{Ramp.} The Brex comparison page says Capital One's acquisition of Brex is ``expected to close in mid-2026'', but it closed on April 7, 2026 \citep{capitalone2026brexclose} (\href{https://ramp.com/versus/brex}{Ramp vs.\ Brex}).
  \item \textbf{Canva.} Section 12 of the Terms of Use is set as body text, while the sections around it are headings (\href{https://www.canva.com/policies/terms-of-use/}{terms of use}).
  \item \textbf{American Airlines.} The tarmac delay plan splits U.S.\ Customs and Border Protection into two list items, so it reads as two agencies (\href{https://www.aa.com/web/i18n/customer-service/support/contingency-plans.html}{contingency plans}).
\end{itemize}

\section{Extra results}
\label{app:extra}

\begin{table*}[t]
  \centering
  \small
  \begin{tabular}{@{}llp{0.62\textwidth}@{}}
    \toprule
    Model & Habit & Measure \\
    \midrule
    Astra & Checks facts as it goes & \numPersFilingWebAstra\% of its turns that record a finding also carry a web call (other models \numPersFilingWebOthersMin\% to \numPersFilingWebOthersMax\%). \numPersNaturalHeldAstra\% of the unplanted factual errors it reports hold up (other models \numPersNaturalHeldOthers\%). \\
    Argon & Searches before recording & \numPersWebBeforeFirstArgon\% of its web calls come before its first finding. \numPersDiagBlockArgon\% of its diagnoses name the block ID (other models at most \numPersDiagBlockOthersMax\%). \\
    Opus & Records early & \numPersFiledByCheckpointOpus\% of its findings are recorded before its context first fills, the most of any model. \\
    Fable & Probes, then types URLs & \numPersProbeFable\% of its audits open with a probe read of each page. \numPersTypedUrlFable\% of its fetches go to addresses no search returned. \\
    Sol & Records without re-reading & \numPersFilingNoReadSol\% of its turns that record a finding carry no read or search. It corrects \numPersRepeatSol\% of its catches in all three runs, the lowest of any model. \\
    Grok & Searches with operators that fail & \numPersOrEmptyGrok\% of its searches chained with OR come back empty. \numPersLimitEndGrok\% of its audits end on a budget limit. \\
    Wonder & Opens the least HTML & \numPersHtmlWonder{} HTML reads per audit, against \numPersHtmlArgon{} for Argon. It misses all six Apple heading issues in every run. \\
    \bottomrule
  \end{tabular}
  \caption{Each model's most distinctive audit habit, measured over the traces of all its audits.}
  \label{tab:habits}
\end{table*}

\paragraph{How each model audits.}
We read the full trace of every run, \numPersRunsRead{} audits across three runs, and \Cref{tab:habits} gives each model's most distinctive habit with the measures behind it.

\begin{figure}[H]
  \centering
  \includegraphics[width=\linewidth]{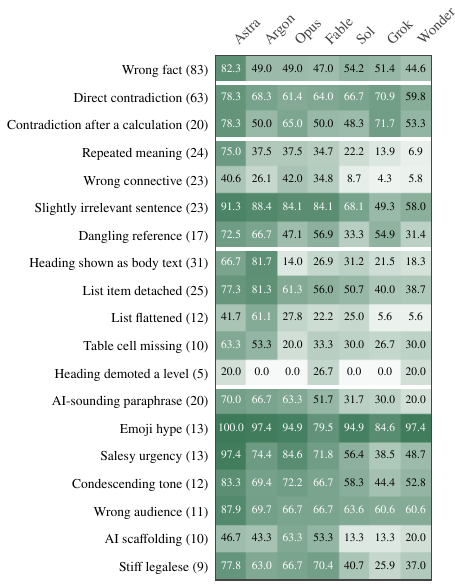}
  \caption{No model corrects more than 42.0\% of wrong connectives. Share (\%) of introduced issues detected and corrected, by kind and model, pooled over runs and brands, since a kind can have a single issue on a brand. The number of introduced issues of each kind is in parentheses, and white rules separate the families.}
  \label{fig:kind-recall}
\end{figure}

\paragraph{Results by kind.}
\Cref{fig:kind-recall} gives the share of each kind of introduced issue that each model detected and corrected.

\paragraph{The logic between sentences is a blind spot.}
No model corrects more than \numConnectiveMax\% of wrong connectives, such as a ``nevertheless'' that reverses the logic, and every model but Astra corrects at most \numRepetitionOthersMaxPct\% of repeated meanings (\Cref{fig:kind-recall}).
Every sentence in these passages reads fine on its own, and the issue shows only in how one sentence relates to the next.

\paragraph{How agents reach a corrected fact.}
For the \numSearchFixes{} wrong facts the frontier models corrected with the web on, we tie each web search and fetch to a fact by its wording and timing and label where the corrected fact first appears (\Cref{tab:hops}).
The agent typed it into a search before any result showed it in \numSearchKnewPct\% of fixes, needed no web call in \numSearchNoWebPct\%, found it in the first tied result in \numSearchOneHopPct\%, and needed several hops in under 1\%.
A hand check of 15 fixes agreed with \numSearchHandCheckPct\% of the labels.

\begin{table}[H]
  \centering
  \small
  \begin{tabular}{@{}lr@{}}
    \toprule
    Where the corrected fact first appears & Fixes (\%) \\
    \midrule
    Typed into a search before any result showed it & \textbf{\numSearchKnewPct} \\
    No web call & \textbf{\numSearchNoWebPct} \\
    First tied result & \numSearchOneHopPct \\
    Several hops & \numSearchMultiHopPct \\
    Unclear & \numSearchUnclearPct \\
    \bottomrule
  \end{tabular}
  \caption{Models mostly know the corrected fact before searching: 79.4\% of corrections came before any search result showed it or with no web call. Where the corrected fact first appears for the \numSearchFixes{} wrong facts the frontier models corrected with the web on, over all three runs. Bold marks the two cases where the agent knew the fact before any search result showed it.}
  \label{tab:hops}
\end{table}

\paragraph{Turn budget.}
\label{app:turns}
We run Sol and Opus with 96 turns instead of 48 over all brands, with every other limit doubled too (\Cref{tab:turns}).
Sol uses the extra turns, taking up to 89, and its corrected share rises from 46.6\% to 52.5\%.
Opus mostly stops well short of the larger budget, taking at most 55 turns, and rises less, from 53.6\% to 56.2\%.
\begin{table}[H]
  \centering
  \small
  \resizebox{\columnwidth}{!}{
  \begin{tabular}{@{}lrrrrrr@{}}
    \toprule
     & \multicolumn{3}{c}{48 turns (default)} & \multicolumn{3}{c}{96 turns} \\
    \cmidrule(lr){2-4}\cmidrule(l){5-7}
    Model & \makecell[r]{Max\\used} & \makecell[r]{Detected\\(\%)} & \makecell[r]{Corrected\\(\%)} & \makecell[r]{Max\\used} & \makecell[r]{Detected\\(\%)} & \makecell[r]{Corrected\\(\%)} \\
    \midrule
    Sol & 48 & 51.4 & 46.6 & 89 & \textbf{56.0} & \textbf{52.5} \\
    Opus & 45 & 61.0 & 53.6 & 55 & \textbf{63.1} & \textbf{56.2} \\
    \bottomrule
  \end{tabular}}
  \caption{Doubling the turns helps Sol, which uses them, more than Opus, which stops at 55. Turn-budget ablation. Max used is the most turns any single audit took under each budget. Detected and corrected are shares of introduced issues (\%), averaged over brands; higher is better, and bold marks the better budget.}
  \label{tab:turns}
\end{table}

\end{document}